\documentclass[11pt,a4paper]{article}
\usepackage[hyperref]{emnlp2020}
\usepackage{times}
\usepackage{latexsym}

\usepackage{times}
\usepackage{latexsym}

\usepackage{url}

\usepackage{graphicx}
\graphicspath{ {pic/} }
\usepackage{amsmath,amssymb}
\usepackage{eucal}
\usepackage{subcaption}
\usepackage[ruled]{algorithm2e}
\usepackage{url}
\usepackage{multirow}
\definecolor{blue}{RGB}{59,118,175}
\definecolor{orange}{RGB}{243,166,103}
\usepackage{microtype}

\aclfinalcopy % Uncomment this line for the final submission
\def\aclpaperid{3031} %  Enter the acl Paper ID here

\newcommand{\tabincell}[2]{\begin{tabular}{@{}#1@{}}#2\end{tabular}}

\title{CREDIT: Coarse-to-Fine Sequence Generation for Dialogue State Tracking}

\author{Zhi Chen, Lu Chen, Zihan Xu, Yanbin Zhao,Su Zhu and Kai Yu \\
  Key Lab. of Shanghai Education Commission for Intelligent Interaction and Cognitive Eng. \\
  SpeechLab, Department of Computer Science and Engineering \\
  Brain Science and Technology Research Center \\
  Shanghai Jiao Tong University, Shanghai, China \\
  {\tt \{zhenchi713, zihan.xu,chenlusz,zhaoyb,paul2204,kai.yu\}@sjtu.edu.cn} }

\begin{document}
\maketitle
\begin{abstract}
% In an end-to-end dialogue system, the dialogue state tracker aims to accurately find a compact representation of the current dialogue status, based on the entire dialogue history. However, the existing approaches regard dialogue states as a combination of different separated triples ({\em domain-slot-value}). In this paper, we reformulate dialogue state tracking as a sequence generation problem with structured state representation. Based on this, the paper proposes to solve it using a {\bf C}oa{\bf R}s{\bf E}-to-fine {\bf DI}alogue state {\bf T}racking ({\bf CREDIT}) approach. Taking advantage of the structured state representation, which is a kind of language sequence, we can further fine-tune the pre-trained model (by supervised learning) through optimizing natural language metrics with the policy gradient method. Like all the generative state tracking methods, CREDIT does not rely on predefined dialogue ontology enumerating all the possible slot values. Empirical results demonstrate our tracker achieves encouraging joint goal accuracy for the five domains in MultiWOZ 2.0 and MultiWOZ 2.1 datasets.

In dialogue systems, a dialogue state tracker aims to accurately find a compact representation of the current dialogue status, based on the entire dialogue history. While previous approaches often define dialogue states as a combination of separate triples ({\em domain-slot-value}), in this paper, we employ a structured state representation and cast dialogue state tracking as a sequence generation problem. Based on this new formulation, we propose a {\bf C}oa{\bf R}s{\bf E}-to-fine {\bf DI}alogue state {\bf T}racking ({\bf CREDIT}) approach. Taking advantage of the structured state representation, which is a marked language sequence, we can further fine-tune the pre-trained model (by supervised learning) by optimizing natural language metrics with the policy gradient method. Like all generative state tracking methods, CREDIT does not rely on pre-defined dialogue ontology enumerating all possible slot values. Experiments demonstrate our tracker achieves encouraging joint goal accuracy for the five domains in MultiWOZ 2.0 and MultiWOZ 2.1 datasets.
\end{abstract}

%  Our model is composed of a hierarchical utterance encoder, a sketch decoder, a sketch encoder, and a state decoder.

\section{Introduction}
Dialogue state tracking (DST) is a core component of the task-oriented dialogue system, which is used to estimate the compact representation of the dialogue context between dialogue agent and user. This compact representation is called the dialogue state, which includes the slot values of different dialogue domains that the user has confirmed. In other words, the dialogue state is a set of {\em domain-slot-value} triples, e.g., {\em $<$hotel$>$-$<$area$>$-centre}. The dialogue agent decides how to respond to the user with this compact representation, based on the dialogue state. In the traditional dialogue system, the upstream component of the dialogue state tracking model is natural language understanding (NLU)~\cite{chen2013unsupervised, zhu2014semantic}. The dialogue state is updated upon the semantic representation of the user's response at the current dialogue turn, parsed by NLU. Nowadays, the end-to-end state tracking approaches~\cite{mrkvsic2016neural, perez2017dialog, ren2018towards} get more and more attention in the DST community, which combine NLU with traditional DST~\cite{henderson2014second, sun2014sjtu} and directly predict the dialogue state based on the dialogue utterances.

\begin{figure}%[htbp!]
\centering
\includegraphics[width=0.5\textwidth]{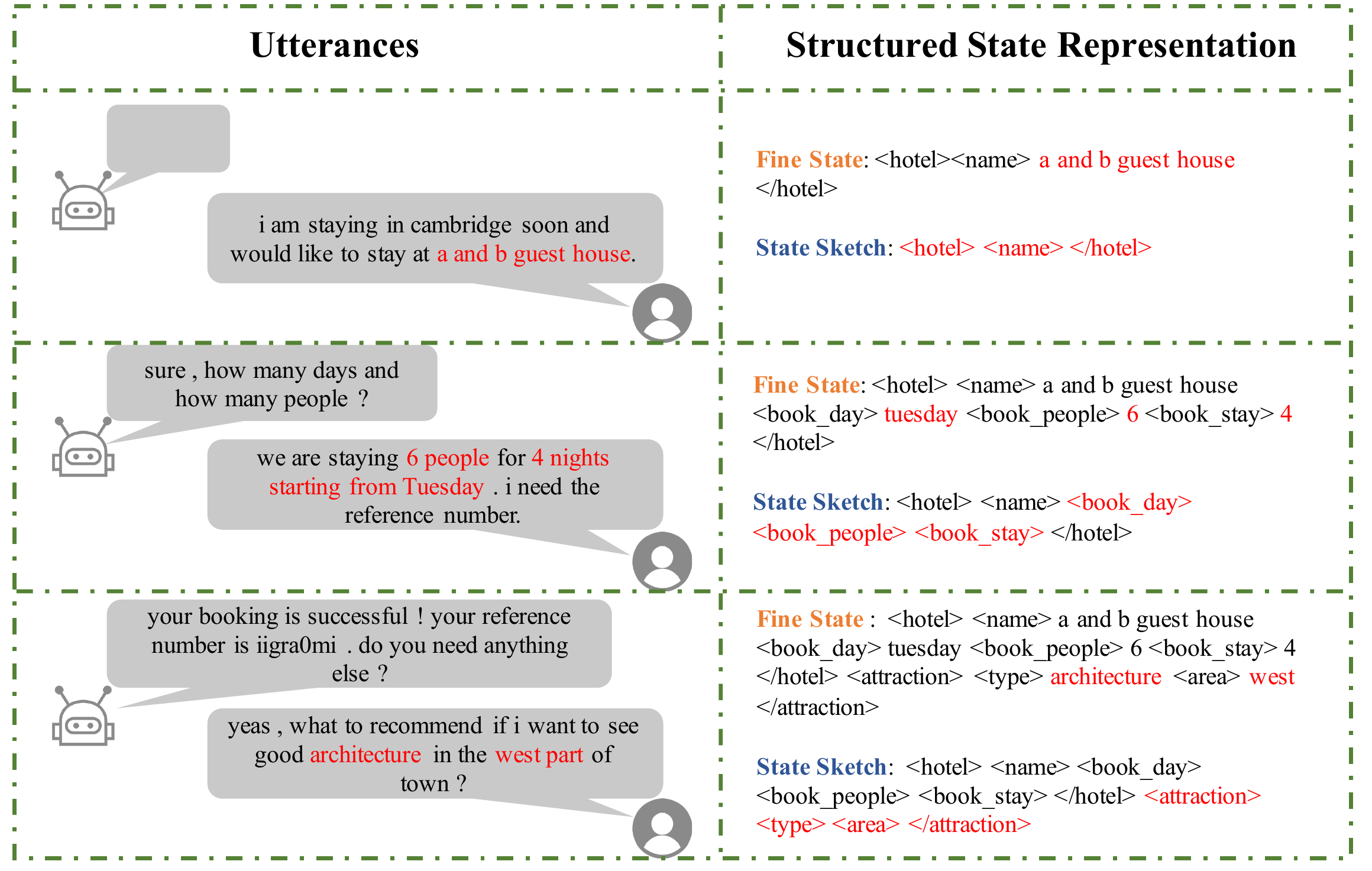}
\caption{The structured state representation and the sketch of the dialogue state.}
\label{fig:example}
\vspace{-6mm}
\end{figure}

Recent proposed end-to-end state tracking approaches can be divided into two categories: classification~\cite{eric2019multiwoz} and generation~\cite{wu-2019-transferable}. The classification methods usually require that all the possible slot values are given by ontology. However, in real dialogue scenarios, some slot values cannot be enumerated. Then the generation methods, which directly extract the slot values from the dialogue history, have been proposed to alleviate this problem. Like the classification methods, most of the generation methods predict slot value one by one, until all the slots on different domains have been visited. The computation cost is proportional to the numbers of the slots. When the dialogue task becomes complex, this problem would be severe.

In this paper, we redefine the dialogue state as structured state representation, as shown in Fig.~\ref{fig:example}. We concatenate all the state triples {\em domain-slot-value} together and use $<${\em /domain}$>$ as a domain separator to cluster all the slot-value pairs of the corresponding domain together. For traditional generation methods~\cite{wu-2019-transferable}, {\em domain-slot} is part of the input, and the entire dialogue state is generated slot by slot. Instead, we regard {\em domain-slot} as the predicted information to decrease the inference complexity. In this paper, the proposed CREDIT is an encoder-decoder model that maps dialogue utterances into structured state representation. Instead of directly generating a complete state, CREDIT first predicts a coarse state representation as an intermediate result, which only contains {\em domain-slot} information and is called {\bf state sketch}. Then, the final state is reasoned from the state sketch and utterances, which is called {\bf fine state}. With cross-entropy loss in supervised learning, the evaluation metric (joint goal accuracy) of the DST method cannot be optimized directly. We adopt the reinforcement learning method to optimize the joint goal accuracy due to the entire dialogue state that is directly generated. Instead of using the accuracy as the reward function, we use smoother BLEU score~\cite{Ranzato2015SequenceLT} as the reward function in our experiments, which will be explained latter. By contrast, NABC~\cite{jurvcivcek2011natural} optimizes the parameters of the tracker with the policy gradient method~\cite{sutton2000policy} under Partially Observable Markov Decision Process (POMDP) framework. Its reward signal is from the user's feedback. However, we optimize the tracking model using the DST metric. To the best of our knowledge, this is the first work to optimize the neural-based state tracking model with the reinforcement learning method.

% Due to the structured state representation that is a sequence, the language metrics (e.g., BLEU score~\cite{Ranzato2015SequenceLT}) can be used to measure the estimated results of CREDIT. With cross-entropy loss in supervised learning, the language metrics of structured state cannot be optimized directly. In this paper, we further leverage reinforcement learning method to fine-tune the pretrained tracking model. To the best of our knowledge, this is the first work to optimize the end-to-end state tracking model with reinforcement learning method.

One-step dialogue state generation and the boost of the performance are the main advantages of CREDIT. The contribution of this paper lies in three parts:
\begin{itemize}
    \vspace{-2mm}
    \item In this paper, we adopt the coarse-to-fine decoding method to generate a structured state. First, the state sketch that consists of domain-slot pairs is decoded. The fine state is produced based on the state sketch and dialogue utterances using copy mechanism~\cite{vinyals2015pointer, wu2016google}.
    \vspace{-3mm}
    \item Our proposed CREDIT is a one-step generation method that does not need to traverse all the possible slots to generate the corresponding dialogue state. CREDIT directly maps dialogue utterances to a structured state, and the inference time complexity (ITC)~\cite{ren2019scalable} is O(1).
    \vspace{-3mm}
    \item To improve CREDIT's final performance, we regard the DST metric as the reward and continue to fine-tune the parameters of the pre-trained tracking model with reinforcement learning loss.
\end{itemize}

\section{Related Work}
\label{sec:relwork}
\paragraph{Multi-Domain DST:}
With the release of the MultiWOZ dataset~\cite{budzianowski2018multiwoz, eric2019multiwoz}, one of the largest task-oriented dialogue datasets, many advanced DST methods for the multi-domain task have been proposed. FJST and HJST ~\cite{eric2019multiwoz} are two straightforward methods. They directly encode the dialogue history into a vector and predict all the slot values by it. Instead of directly concatenating the whole dialogue history as input in FJST, HJST takes the hierarchical model as the encoder. HyST~\cite{goel2019hyst} is a hybrid method that improves HJST by adding the value-copy mechanism. With the advance of the pre-trained models (e.g., BERT~\cite{devlin-2019-bert}), SUMBT~\cite{lee2019sumbt} first uses pre-trained BERT to encode the dialogue utterances and a slot into a vector and computes the distances with the candidate values of the corresponding slot. TRADE~\cite{wu-2019-transferable} is the first work that directly generates the slot value from the dialogue history. Following TRADE, there are some improved versions, DS-DST~\cite{zhang2019find}, DST-picklist~\cite{zhang2019find}, and SOM-DST~\cite{kim2019efficient}. DS-DST and DST-picklist are proposed simultaneously, which divide the slots as the uncountable type and countable type and generate the slot value in a hybrid method like HyST. Compared with DS-DST, DST-picklist knows all the candidate values of the slots, including uncountable slots. Unlike TRADE, SOM-DST feeds dialogue history and the previous state as the input and modifies the state with dialogue history into the current state. These improved versions rely on the pre-trained BERT as the utterance encoder. Different from the above methods, DST Reader~\cite{gao2019dialog} formulates the DST task as a machine reading task and leverages the corresponding method to solve the multi-domain task. 

In this work, we formulate the DST task as a structured state generation task and solve it using the coarse-to-fine decoding method. COMER~\cite{ren2019scalable} is the closest method to our proposed method CREDIT that directly generates all the domain-slot-value triples. However, the inputs of COMER include the previous predicted state and utterances of the current turn, and the generated state is tree-like instead of a structured state sequence. 

\begin{figure*}%[htbp!]
\centering
\includegraphics[width=0.8\textwidth]{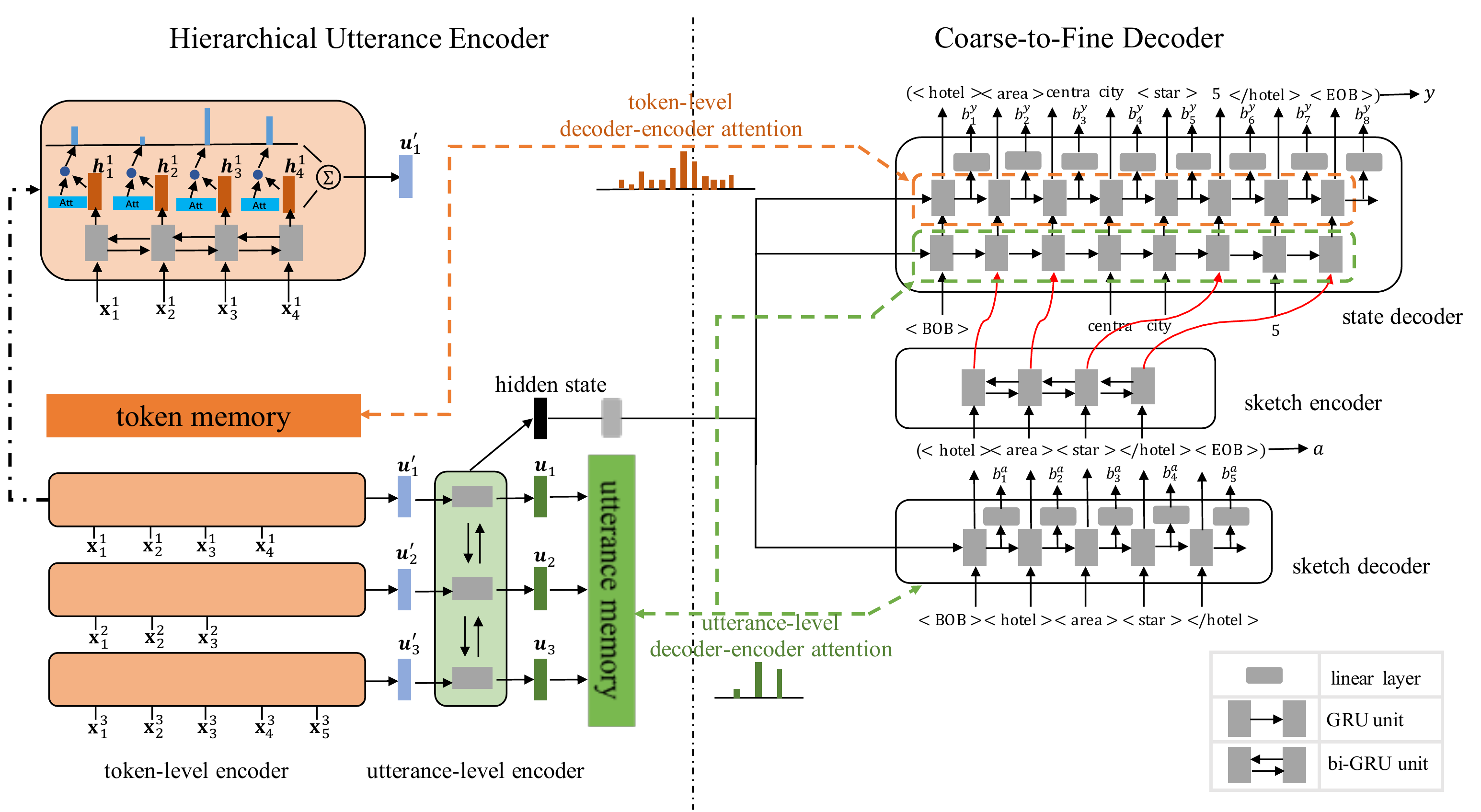}
\caption{The coarse-to-fine state generation model consists of four models: hierarchical utterance encoder, sketch decoder, sketch encoder and state decoder, where the token memory stores token vectors encoded by token-level encoder and the utterance memory collects utterance vector encoded by utterance-level encoder. $b^a_i$ and $b^y_i$ are the baseline values used in RL optimization.}
\label{fig:credit_model}
\vspace{-6mm}
\end{figure*}
\vspace{-2mm}
\paragraph{Hierarchical Generation:}
The coarse-to-fine generation method is adopted by ~\citet{dong2018coarse} for semantic parsing tasks to simplify the generation task of structured state representation. In the DST task, the input is a multi-turn dialogue rather than an utterance in semantic parsing tasks, which makes it hard. Our proposed CREDIT improves the coarse-to-fine decoding method with a two-stage attention mechanism. Recently, \citet{liu2019automatic} leverages the hierarchical generation method to solve the dialogue summary task, which first generates the predefined dialogue summary actions and then generates the final summary. 

\section{CREDIT Model}
\label{sec:credit}
The proposed model consists of four components: a hierarchical utterance encoder, a state sketch decoder, a sketch encoder, and a state decoder. Instead of predicting all the slot values, we generate the dialogue state, which does not include ``none'' value~\cite{wu-2019-transferable} of the slot, as shown in Fig.~\ref{fig:example}. We redefine the dialogue state as a structured representation~\cite{tang2000automated,andreas-2013-semantic,zhao2015type}. One of the main challenges is that the DST task needs to reason among all the utterances of the whole dialogue to get the complete dialogue state.

The hierarchical utterance encoder comprises two different encoders: a token-level encoder and an utterance-level encoder. The token-level encoder encodes each token of the utterances into a token vector. The utterance-level encoder encodes each utterance of a dialogue into an utterance vector using the token vectors with attention mechanism~\cite{luong-2015-effective}. The state sketch decoder predicts a sketch of a dialogue state that consists of the mentioned domain-slot pairs in a dialogue, as shown in Fig.~\ref{fig:example}. Since there is no need to copy any values from the utterances for the sketch decoder, its decoder-encoder attention is only accomplished with the utterance vectors. The sketch encoder encodes the predicted state sketch into a sequence of sketch vectors. Different from sketch decoder, the specific slot values could be directly copied from the dialogue utterances for the state decoder. There are two-level decoder-encoder attentions in the state decoder model: utterance-level attention with the utterance vectors (store in utterance memory) and token-level attention (store in token memory) with the token vectors. Apart from using the embedding of the previous token, the state decoder is fed with the corresponding sketch vector, when the previous token exists in state sketch.

Let us define $x=\{x^1, x^2, \dots, x^{|x|}\}$ as the set of dialogue utterances and $y=\{y_1, y_2, \dots, y_{|y|}\}$ as its structured state representation. Our goal is to learn state tracker from the dialogue utterances $x$ with structured state representation $y$. We estimate $p(y|x)$, the conditional probability of state representation $y$ given $x$. Under coarse-to-fine formulation, we decompose $p(y|x)$ into a two-stage generation process:
\vspace{-3mm}
\begin{align}
p(y|x) = p(y|x, a)p(a|x),
\end{align}
where $a=\{a_{1}, a_{2}, \dots, a_{|a|}\}$ is state sketch representing the abstraction of $y$.
\vspace{-2mm}
\subsection{Hierarchical Utterance Encoder}
Hierarchical utterance encoder encodes an utterance into a sequence of token vectors and a contextual utterance vector.
\vspace{-2mm}
\paragraph{Token-level encoder:}
We use a bi-directional gate recurrent units (GRU) to encode an utterance. The input of token-level encoder is an utterance of a dialogue, represented as $x^i = \{\mathbf{x}^i_{1}, \mathbf{x}^i_{2}, \dots, \mathbf{x}^i_{|x^i|}\}$, where $\mathbf{x}^i_{j}$ means the $j$-th token embedding of $i$-th dialogue utterance and $d_{emb}$ indicates the embedding size. The encoded utterance is represented as $\mathbf{h}^i = \{\mathbf{h}^i_{1}, \mathbf{h}^i_{2}, \dots, \mathbf{h}^i_{|x^i|}\}$ via:
\vspace{-4mm}
\begin{align}
\overrightarrow{\mathbf{h}}^i_j &= f^{x}_{\text{GRU}}(\mathbf{h}^i_{j-1}, \mathbf{x}^i_j), j=1,\dots, |x^i|, \notag \\
\overleftarrow{\mathbf{h}}^i_j &= f^{x}_{\text{GRU}}(\mathbf{h}^i_{j+1}, \mathbf{x}^i_j), j=|x^i|,\dots, 1, \notag \\
\mathbf{h}^i_j &= [\overrightarrow{\mathbf{h}}^i_j, \overleftarrow{\mathbf{h}}^i_j] \notag
\end{align}
where $[\cdot, \cdot]$ means the concatenation of two vectors and $f^{x}_{\text{GRU}}$ is the token-level bi-directional GRU function. 

\paragraph{Utterance-level encoder:}
We use attention mechanism to make encoder module pay more attention to state-related tokens. For each token $x^i_j$ in utterance $x^i$, the attention score is calculated as:
\begin{align}
\text{score}(\mathbf{x}^i_j) = \mathbf{v}^{\mathsf{T}}\text{tanh}(\mathbf{W}^S(\mathbf{h}^i_j) + \mathbf{b}^S), \notag
\end{align}
where $\mathbf{v}^{\mathsf{T}}$, $\mathbf{W}^S$ and $\mathbf{b}^S$ are trainable parameters. The attentive summarization over $\mathbf{h}^i$ is applied to get a utterance vector:
\vspace{-2mm}
\begin{align}
\mathbf{u}_i^{\prime} = \text{softmax}(\text{score}(\mathbf{x}^i))\mathbf{h}^i. \notag
\end{align}
Here we get utterance-level representation $\mathbf{u}^{\prime} = \{\mathbf{u}_1^{\prime}, \mathbf{u}_2^{\prime}, \dots, \mathbf{u}_{|x|}^{\prime}\}$. Since the token vectors in $\mathbf{h}^i$ do not contain any context information, the utterance vector $\mathbf{u}_i^{\prime}$ is context-free. We feed context-free utterance vector $\mathbf{u}_i^{\prime}$ to another bi-directional GRU to get contextual utterance-level representation $\mathbf{u} = \{\mathbf{u}_1, \mathbf{u}_2, \dots, \mathbf{u}_{|x|}\}$ via:
\begin{align}
\overrightarrow{\mathbf{u}}_i &= f^{u}_{\text{GRU}}(\mathbf{u}_{i-1}, \mathbf{u}_i^{\prime}), i=1,\dots, |x|, \notag\\
\overleftarrow{\mathbf{u}}_i &= f^{u}_{\text{GRU}}(\mathbf{u}_{i+1}, \mathbf{u}_i^{\prime}), i=|x|,\dots, 1, \notag\\
\mathbf{u}_i &= [\overrightarrow{\mathbf{u}}_i, \overleftarrow{\mathbf{u}}_i],
\label{eq:bi_gru}
\end{align}
where $f^{u}_{\text{GRU}}$ is the utterance-level bi-directional GRU function.

\subsection{Coarse Sketch Decoder}
\label{sect:skt_dec}
The sketch decoder learns to compute $p(a|x)$ and generates a state sketch $a$ conditioned on dialogue context. We use a GRU to decode state sketch. At $t$-th time step of sketch decoding, the decoder's hidden vector is computed by $\mathbf{d}_t^a = f_{\text{GRU}}^{a}(\mathbf{d}_{t-1}^a, \mathbf{a}_{t-1})$, where $\mathbf{a}_{t-1}$ is the embedding of the previously predicted token and $f_{\text{GRU}}^{a}$ is the sketch decoding GRU function. The hidden state of the first time step in the decoder is initialized by the last hidden state of $f_{\text{GRU}}^{u}$ through a linear layer. Additionally, we use decoder-encoder attention mechanism to learn soft alignment. At the current time step $t$ of the decoder, we compute the attention score with $i$-th utterance-level representation $\mathbf{u}_i$ as:
\begin{align}
s_{t,i}^{a2u} = \text{exp}(\mathbf{d}_t^a \cdot \mathbf{u}_i)/Z_t,
\label{eq:att1}
\end{align}
where $Z_t=\sum_{j=1}^{|x|}\text{exp}(\mathbf{d}_t^a \cdot \mathbf{u}_j)$ is a normalization term. Then we calculate $p(a_t|a_{<t}, x)$ via:
\begin{align}
\label{eq:att2}
\mathbf{e}_t^{a2u} &= \sum_{i=1}^{|x|}s_{t,i}^{a2u}\mathbf{u}_i,  \\
\mathbf{d}_t^{a2u} &= \text{tanh}(\mathbf{W}_1[\mathbf{d}_t^a, \mathbf{e}_t^{a2u}] + \mathbf{b}_1),  \notag\\
P^a(a_t|a_{<t}, x) &= \text{softmax}(\mathbf{W}_a \mathbf{d}_t^{a2u} + \mathbf{b}_a), \notag
\end{align}
where $\mathbf{W}_1, \mathbf{b}_1, \mathbf{W}_a, \mathbf{b}_a$ are trainable parameters and $a_{<t}=[a_1, \dots, a_{t-1}]$. Generation terminates until the end token of sequence ``$<$EOB$>$'' is emitted.

\subsection{Fine State Generation}
\label{sect:state_dec}
The structured state is predicted by conditioning on the dialogue context $x$, and the generated state sketch $a$. The model uses encoder-decoder architecture with a two-stage attention mechanism to generate the final state.

\paragraph{State Sketch Encoder:}
As shown in Fig.~\ref{fig:credit_model}, a bi-directional GRU function maps the state sketch into a sequence of sketch vectors $\{\mathbf{g}_t\}_{t=1}^{|a|}$ as Equation~\ref{eq:bi_gru}, where $\mathbf{g}_t$ is $t$-th encoded vector.

\paragraph{Fine State Decoder: }
The final decoder is based on two GRUs with a two-stage attention mechanism, as shown in Fig.~\ref{fig:credit_model}. At the $t$-th time step of the decoder, the hidden vector of the bottom GRU is computed by $\mathbf{d}_t^{y_1} = f_{\text{GRU}}^{y_1}(\mathbf{d}_{t-1}^{y_1}, \mathbf{i}_t)$, where $\mathbf{i}_t$ is 
\begin{equation}
\label{eq:final_input}
\mathbf{i}_t=\left\{
\begin{aligned}
\mathbf{g}_k&, \text{$y_{t-1}$ is equal to $a_k$} \\
\mathbf{y}_{t-1}&, \text{otherwise},
\end{aligned}
\right.
\end{equation}
and $\mathbf{y}_{t-1}$ is the embedding of the previously predicted token. Once the previous output $y_{t-1}$ appears in generated state sketch, we use the corresponding encoded sketch vector as the input of the decoder. We compute the first stage attention score $s_{t,i}^{y2u}$ with utterance-level representation $\mathbf{u}_i$ and the hidden state $\mathbf{d}_t^{y_1}$, and get the weighted-sum attention vector $\mathbf{e}_t^{y2u}$, which is the same as Equation~\ref{eq:att1} and Equation~\ref{eq:att2}. The output of the bottom GRU at the $t$-th time step is 
\begin{align}
\mathbf{d}_t^{y2u} = \text{tanh}(\mathbf{W}_{y_1}[\mathbf{d}_t^{y_1}, \mathbf{e}_t^{y2u}]+ \mathbf{b}_{y_1}), \notag
% \label{eq:y2u}
\end{align}
where $\mathbf{W}_{y_1},  \mathbf{b}_{y_1}$ are parameters.

The input of the top GRU at the $t$-th time step is $\mathbf{d}_t^{y2u}$ and the output is $\mathbf{d}_t^{y_2}=f_{\text{GRU}}^{y_2}(\mathbf{d}_{t-1}^{y_2}, \mathbf{d}_t^{y2u})$. The second stage attention score is calculated with token-level representation $\mathbf{h}$ as
\begin{align}
s_{t,i,j}^{y2t} = s_{t,i}^{y2u}(\text{exp}(\mathbf{d}_t^{y_2} \cdot \mathbf{h}^i_j)/Z_{t,i}),
\label{eq:y2t1}
\end{align}
where $Z_{t,i}=\sum_{j=1}^{|x^i|}\text{exp}(\mathbf{d}_t^{y_2}\cdot \mathbf{h}^i_j)$ is the normalization term and $\mathbf{h}^i_j$ is the $j$-th token of the $i$-th utterance. Then we compute the generation probability $p_g(y_t|y_{<t},x,a)$ via:
\begin{align}
\mathbf{e}_t^{y2k} &= \sum_{i=1}^{|x|}\sum_{j=1}^{|x^i|}s_{t,i,j}^{y2t}\mathbf{h}^i_j, \notag \\
\mathbf{d}_t^{y2k} &= \text{tanh}(\mathbf{W}_{y_2}[\mathbf{d}_t^{y_2}, \mathbf{e}_t^{y2t}, \mathbf{e}_t^{y2u}] + \mathbf{b}_{y_2}), \notag \\
\label{eq:pg}
P_g(y_t|y_{<t}&, x, a) = \text{softmax}(\mathbf{W}_y \mathbf{d}_t^{y2k} + \mathbf{b}_y),
\end{align}
where $\mathbf{W}_{y_2}, \mathbf{b}_{y_2}, \mathbf{W}_y, \mathbf{b}_y$ are parameters. The final output distribution is the weighted-sum of two different distributions,
\begin{align}
\label{eq:final}
P^y(y_t|y_{<t}, x, a) = p^{gen}_t P_g + (1-p^{gen}_t)P_c,
\end{align}
where the distribution $P_c$ comprises the second stage attention score $s_{t,i,j}^{y2t}$ of the decoder. The scalar $p^{gen}_t$ is computed by
\begin{align}
p^{gen}_t = \text{sigmoid}(\mathbf{W}_{gen}[\mathbf{d}_t^{y_2}, \mathbf{e}_t^{y2t}, \mathbf{e}_t^{y2u}] + \mathbf{b}_{gen}), \notag
\end{align}
where $\mathbf{W}_{gen}, \mathbf{b}_{gen}$ are parameters.

\subsection{Inference}
Because the generated state has to be translated into domain-slot-value triples at test time, the greedy decoding method is not suitable in the structured state generation task. In this work, we adopt the masked greedy decoding method at test time. For example, in a structured state, the first token must be a domain or the state ending token. We design some handcraft masking rules to ensure that the final generated state is translatable.

\section{Learning Objective}
\label{sec:lobj}
In this paper, we use both the cross-entropy loss and reinforcement loss. Previous works mainly use only cross-entropy loss. Since we reformat the dialogue state as a sequence, the BLEU score can be used to evaluate the predicted dialogue state. However, there is a gap between cross-entropy loss and the metric BLEU. So we consider the BLEU score as a reinforcement reward and optimize it directly.

\paragraph{Cross-entropy Loss:}
For sequence generation tasks, the most widely used method to train an encoder-decoder model is minimizing a cross-entropy loss at each decoding step. This method is called the teacher-forcing algorithm~\cite{williams1989learning}. For state sketch decoder, the reference of the given dialogue is $\{\hat{a}_1, \dots, \hat{a}_m\}$. Its cross-entropy loss is 
\vspace{-4mm}
\begin{align}
\label{eq:ce_a}
L_{ce}^a = \frac{1}{m}\sum_{i=1}^m P^a(\hat{a}_i),
\end{align}
where $P^a(\hat{a}_i)$ is the predicted probability of $\hat{a}_i$. Similarly, for state decoder, the reference of the given dialogue is $\{\hat{y}_1, \dots, \hat{y}_n\}$. The corresponding cross-entropy loss is
\vspace{-4mm}
\begin{align}
\label{eq:ce_y}
L_{ce}^y = \frac{1}{n}\sum_{i=1}^n P^y(\hat{y}_i),
\end{align}
where $P^y(\hat{y}_i)$ is the predicted probability of $\hat{y}_i$.

\paragraph{Reinforcement Loss:}
Minimizing the cross-entropy can not guarantee to improve the evaluation metric (joint accuracy). For generation tasks, there are exposure bias problem~\cite{Ranzato2015SequenceLT}. During the training procedure with cross-entropy loss, the previous token is normally from the reference. However, when predicting the sequence, the input token is sampled by the model instead of the reference. So the error could accumulate in prediction. A good choice is to adopt the policy gradient method (one of the RL methods) to alleviate this problem by using the metric as the reward function. In our experiment, we use the BLEU score as the reward function rather than joint accuracy. The main reason is that the joint accuracy reward will cause a serious sparse reward problem during RL training. For example, the truth state is ``{\em $<$hotel$>$ $<$name$>$ a and b guest house $<$area$>$ centre $<$/hotel$>$}''. The joint accuracy rewards of these two predicted states ``{\em $<$hotel$>$ $<$name$>$ a and b guest $<$area$>$ centre $<$/hotel$>$}'' and ``{\em $<$hotel$>$ $<$name$>$ a a a a $<$area$>$ centre $<$/hotel$>$}'' are the same, where only one slot is correctly predicted. However, the first prediction is apparently better than the second. This sparse reward problem leads the training process unstable. By contrast, the BLEU score exactly indicates the degree of error of the predicted sequence, where the less the overlap between our generated state and ground truth is, the lower the BLEU score is.

From the sketch generation view of RL, we treat the generated state sketch as an action sequence. For the sketch sequence decoding, the sequence $\overline{a}=\{\overline{a}_1, \dots, \overline{a}_m\}$ is sampled from $P^a$ at each decoding time step. The reward function is defined as below:
\vspace{-2mm}
% \begin{small}
\begin{equation}
R^a(\overline{a}_i)=\left\{
\begin{aligned}
\text{BLEU}(\overline{a}, \hat{a})&, \text{$\overline{a}_i=$ $<$EOB$>$} \\
-0.01&, \text{$\overline{a}_i \notin \hat{a}$} \\
0.01&, \text{$\overline{a}_i \in \hat{a}$},
\end{aligned}
\right.
\notag
\end{equation}
% \end{small}
where BLEU$(\cdot,\cdot)$ is BLEU function and $\hat{a}$ is the ground truth. After implementing a sketch decoding, we compute the discount accumulated reward for each decoding action by
% \begin{small}
\vspace{-2mm}
\begin{align}
% \label{eq:return}
r^a(\overline{a}_i) = R^a(\overline{a}_i) + \lambda R^a(\overline{a}_{i+1}) + \dots + \lambda^{m-i} R^a(\overline{a}_{m}), \notag
\end{align}
% \end{small}
where $\lambda$ is discount rate. In order to stabilize the RL process, the baseline value $b^a(\overline{a}_i)$ is estimated by 
% \begin{small}
\vspace{-2mm}
\begin{align}
% \label{eq:baseline}
b^a(\overline{a}_i) = \mathbf{W}^a_{bse}\mathbf{d}_i^{a} + \mathbf{b}^a_{bse}, \notag
\end{align}
% \end{small}
where $\mathbf{W}^a_{bse}, \mathbf{b}^a_{bse}$ are parameters and $\mathbf{d}_i^{a}$ is $i$-th hidden state of sketch decoding GRU in sketch decoder introduced in Section~\ref{sect:skt_dec}. According to the Policy Gradient therom~\cite{sutton2000policy}, the RL loss is 
% \begin{small}
\vspace{-2mm}
\begin{align}
\label{eq:rl_a}
L_{rl}^a = -\frac{1}{m}\sum_{i=1}^m (r^a(\overline{a}_i)-b^a(\overline{a}_i))^2\log(P^a(\overline{a}_i)).
\end{align}
% \end{small}

Similarly, we define the RL loss for a fine state as:
% \begin{small}
\vspace{-2mm}
\begin{align}
\label{eq:rl_y}
L_{rl}^y = -\frac{1}{n}\sum_{i=1}^n (r^y&(\overline{y}_i)-b^y(\overline{y}_i))^2\log(P^y(\overline{y}_i)), \\
b^y(\overline{y}_i) &= \mathbf{W}^y_{bse}\mathbf{d}_i^{y_2} + \mathbf{b}^y_{bse}, \notag
\end{align}
% \end{small}
where $\overline{y}_i$ is $i$-th token of predicted state $\overline{y}$, $r^y$ is accumulated reward function that is the same as $r^a$, $b^y$ is baseline function for state decoding and $\mathbf{W}^y_{bse}$, $\mathbf{b}^y_{bse}$ are parameters.

Finally, we combine the Equations~\ref{eq:ce_a},~\ref{eq:ce_y},~\ref{eq:rl_a},~\ref{eq:rl_y} to get the final loss:
\begin{small}
\begin{align}
% \label{eq:loss}
L = \beta_1L_{ce}^a + \beta_2L_{ce}^y + \beta_3L_{rl}^a + (1-\beta_1-\beta_2-\beta_3)L_{rl}^y, \notag
\end{align}
\end{small}
where $\beta_1, \beta_2, \beta_3$ are hyper-parameters.

\section{Experiments}
\subsection{Datasets}
\paragraph{MultiWOZ 2.0:} 
MultiWOZ is the largest task-oriented dialogue dataset for multi-domain dialogue state tracking, which contains 8438 multi-turn dialogues and spans 7 dialogue domains. For the DST task, there are only five domains (\emph{restaurant, hotel, attraction, taxi, train}) in the validation and test set. The domains \emph{hospital, bus} only exist in the training set.

\begin{small}
\begin{table*}[h]%!hbp
\begin{center}
\begin{tabular}{ c|c|c|c|c|c }
%  \hline
%  \multicolumn{4}{|c|}{Country List} \\ \tabincell{c}{Joint Acc.\\MultiWOZ 2.0}
 \hline
Model & +BERT & \tabincell{c}{Joint Acc.\\MultiWOZ 2.0} & \tabincell{c}{Joint Acc.\\MultiWOZ 2.1}  & ITC & Model Size \\
 \hline
SUMBT~\cite{lee2019sumbt} & Y  & 42.40\% & -  &   O(MN) & $>$110M \\
COMER~\cite{ren2019scalable}  & Y & 48.79\% & -   &   O(1) & $>$340M \\
DS-DST~\cite{zhang2019find} & Y  & - & 51.21\%   &   O(M) & $>$110M \\
DST-picklist~\cite{zhang2019find} & Y  & - & \bf{53.30\%}   &   O(MN) & $>$110M \\
SOM-DST~\cite{kim2019efficient} & Y  & \bf{51.72}\% & 53.01\%   &   O(1) & $>$110M \\
 \hline
$\text{HJST}^*$~\cite{eric2019multiwoz} & N  & 38.40\% & 35.55\%   &   O(M) & - \\
$\text{DST Reader}^*$~\cite{gao2019dialog} & N  & 39.41\% & 36.40\%  &   O(M) & - \\
$\text{FJST}^*$~\cite{eric2019multiwoz} & N  & 40.20\% & 38.00\%  &   O(M) & - \\
$\text{HyST}^*$~\cite{goel2019hyst}  & N & 42.33\% & 38.10\%  &   O(M) & - \\
TRADE~\cite{wu-2019-transferable} & N  & 48.60\% & 45.60\%  &   O(M) & 10M \\
CREDIT-SL(ours) & N  & 51.18\% & 50.11\%  &   O(1) & 44M \\
CREDIT-RL(ours) & N  & \bf{51.68\%} & \bf{50.61\%}  &   O(1) & 44M \\
 \hline
\end{tabular}
\caption{The results of baseline models and our proposed CREDIT on MultiWOZ 2.0 and MultiWOZ 2.1 dataset. +BERT means that the tracking model encodes the utterances using pretrained BERT. In ITC column, $M$ is the number of slots and $N$ is the number of values. $*$ indicates the model has not released its source code and the model size is unknown. Joint Acc. means the joint goal accuracy.}
\label{tab:result}
\vspace{-4mm}
\end{center}
\end{table*}
\end{small}

\paragraph{MultiWOZ 2.1:}
MultiWOZ 2.1 fixes the state annotations and utterances in MultiWOZ 2.0. In MultiWOZ 2.0, there are five common error types in dialogue state annotations: delayed markups, multi-annotations, mis-annotations, typos, and forgotten values. The specific introduction of these errors described in ~\cite{eric2019multiwoz}. However, the number of dialogues in MultiWOZ 2.1 is the same as in MultiWOZ 2.0.

\subsection{Evaluation Metrics}
\paragraph{Joint Goal Accuracy:} 
This is the standard metric for dialogue state tracking tasks. Joint goal success of dialogue means that the values of all the domain-slot pairs are correctly predicted. In this work, we adopt the implementation in TRADE~\cite{wu-2019-transferable} for this metric.

\paragraph{BLEU Score:} 
The BLEU score can measure the performance of our model because we redefine the dialogue state as a sequence.

\paragraph{ITC:} 
ITC means the inference time complexity of the model, which is first proposed in ~\cite{ren2019scalable}.

\paragraph{Model Size:} 
In this work, we measure the scale of the tracking models. We notice that recently proposed methods use the pre-trained models (e.g., BERT~\cite{devlin-2019-bert}) as part of the tracking model to promote the performance. In the real scenario, the trade-off between parameter scale and performance is the crucial point.

\subsection{Training Details}
Similar to TRADE, we initialize all the embeddings using the concatenation of Glove embeddings~\cite{pennington2014glove} and character embeddings~\cite{hashimoto2017joint}. We divide the training process into two stages: supervised learning (SL) pretraining stage and reinforcement learning (RL) fine-tuning stage. At the SL stage, we only use the cross-entropy loss to train the model. We adopt Adam optimizer~\cite{kingma2014adam} with learning rate 1e-4. We set the batch size to 16 and dropout rate to 0.2. At the RL stage, We modify the learning rate to 1e-5 and dropout rate to 0. The hyper-parameters in Equation~\ref{eq:final} are $\beta_1=0.25, \beta_2=0.25, \beta_3=0.25$. The discount rate $\lambda$ of RL is 1.

\subsection{Results}
% \begin{figure}%[htbp!]
% \centering
% \includegraphics[width=0.49\textwidth]{fig/acc_bie_dia.pdf}
% \caption{The \textcolor{blue}{blue} curve indicates joint goal accuracy and the \textcolor{orange}{orange} curve indicates joint sketch accuracy. (a) demonstrates the effect of dialogue length on these two kinds of joint accuracy. (b) shows the effect of ground truth length.}
% \label{fig:acc}
% \vspace{-4mm}
% \end{figure}
\paragraph{Joint goal accuracy:}
As shown in Table~\ref{tab:result}, our proposed CREDIT achieves the best performance in the BERT-free models, where CREDIT-SL only uses the cross-entropy loss and CREDIT-RL is fine-tuned by RL method introduced in Section~\ref{sec:lobj} from pre-trained CREDIT-SL. CREDIT-RL can get a further performance boost in MultiWOZ 2.0 and MultiWOZ 2.1 dataset. As mentioned in Section~\ref{sec:relwork}, COMER is the closest method to CREDIT. Compared with COMER, the benefit of CREDIT comes from the structured state representation. It not only contains the structured information of the dialogue state but also is a sequence that is easy to be generated by traditional encoder-decoder models. In COMER, the generated state is a tree where the generated values of different slots are independent. The tree-like state could lead to a serious problem in the DST task. Especially in the MultiWOZ dataset, there are lots of same values in the slot \emph{departure} and \emph{destination}. The departure value is different from destination value and can not be captured by tree-like state representation, but the structured state can.

\begin{figure*}%[htbp!]
\centering
\includegraphics[width=0.9\textwidth]{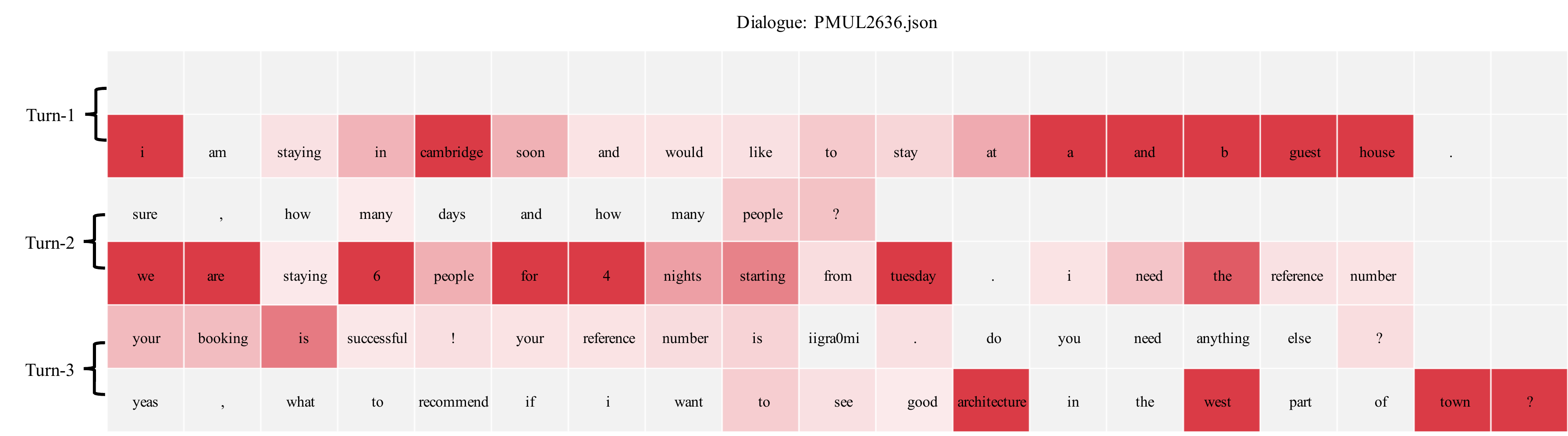}
\caption{The attention heat map of an dialogue example whose dialogue ID is PMUL2636.json from MultiWOZ 2.1 dataset.}
\label{fig:heat}
\vspace{-4mm}
\end{figure*}

\begin{table*}[h]%!hbp
\begin{center}
\begin{tabular}{ l|c|c|c }
%  \hline
%  \multicolumn{4}{|c|}{Country List} \\ \tabincell{c}{Joint Acc.\\MultiWOZ 2.0}
 \hline
 Model & Joint Acc. & Joint Sketch Acc.  & BLEU \\
 \hline
CREDIT-SL & 50.11\% & 57.26\%    & 72.08  \\
(-)Coarse Decoder & 48.62\% & /    &71.83  \\
(-)Copy Mechanism & 47.80\% & 56.63\%    & 71.69  \\
(-)Hierarchical Encoder & 46.82\% & 53.92\%    & 71.54  \\
 \hline
\end{tabular}
\caption{The results of the ablation experiments on MultiWOZ 2.1 dataset. (-) means removing or changing the corresponding component.}
\label{tab:ablation}
\vspace{-6mm}
\end{center}
\end{table*}
\vspace{-2mm}
\paragraph{ITC:} With the number of slots increasing, inference time becomes an essential factor that depends on the reaction speed of the dialogue system. Our proposed CREDIT directly generates all the possible domain-slot-value triples at once and does not need to visit all the slots predefined by the ontology. Thus, the inference time complexity of CREDIT is O(1). Compared with SOM-DST whose ITC is also O(1), its input needs to include all the slot-value pairs of previous dialogue state. In other words, this method decreases the inference time complexity by increasing memory store complexity. However, our proposed CREDIT does not increase inference time complexity and memory store complexity, with the number of the slots increasing.

% \paragraph{Influence analysis:}
% In Fig.~\ref{fig:acc}, we validate the effects of dialogue length and ground truth length. The ground truth length means the number of domain-slot-value triples in a dialogue state. The dialogue length is turn length of the dialogue. Two lines \emph{state} and \emph{sketch} represent joint state accuracy and joint sketch accuracy, respectively. We can see that two kinds of joint accuracy both decrease rapidly when the dialogue length or the state length increases. The descending trend of these two kinds of joint accuracy is completely consistent.
\vspace{-2mm}
\paragraph{Example analysis:}
Fig.~\ref{fig:heat} shows a heat map of a dialogue example chosen in the test set of MultiWOZ 2.1, whose dialogue state is correctly predicted by our CREDIT-RL model. The heat weight of every token in the dialogue is the sum of token-level attention weights computed in fine state decoder by Equation~\ref{eq:y2t1}. As shown in Fig.~\ref{fig:heat} , when generating the dialogue state, our proposed CREDIT is able to focus on the right value-related tokens, like \emph{a and b guest house}, \emph{6 people} and so on.
\vspace{-2mm}
\paragraph{BERT augment:}
We have tried to replace the GRU with the pre-trained BERT in CREDIT's token-level encoder. But its performance is near to BERT-free CREDIT. We think there are two main reasons leading to this situation. The first reason is from the MultiWOZ task itself. The user's goal in each utterance usually is direct and straightforward. The challenge is to model the relationship among different utterances rather than internal utterance. The second reason is from the task definition. We put all the dialogue history as the input. However, the utterances are fed into BERT one by one in the token-level encoder. BERT, whose parameters are pre-trained with the sentence-pair task, is not suitable for this single sentence task.

\subsection{Ablation Experiment}
Table~\ref{tab:ablation} shows the results of ablation studies. We validate the effects of three factors: coarse sketch decoder, copy mechanism used in fine state decoder, and hierarchical utterance encoder.

\vspace{-2mm}
\paragraph{The effect of coarse decoder:}
We remove the coarse sketch decoder and sketch encoder from CREDIT in this validation study. When generating the final state sequence in fine state decoder, we only use the embedding of the previously predicted token $\mathbf{y}_{t-1}$ as input instead of $\mathbf{i}_t$ computed by Equation~\ref{eq:final_input}. There is no sketch information to guide the state generation process. As shown in Table~\ref{tab:ablation}, without the coarse decoder, there is significant performance degradation. It illustrates that the coarse-to-fine generation mechanism can simplify the structured state generation process, and sketch information is helpful to the fine state generation.
\vspace{-2mm}
\paragraph{The effect of copy mechanism:}
In this study, we directly use the generation probability $P_g$ computed by Equation~\ref{eq:pg} as the final output distribution in fine state decoder. In the DST task, the copy mechanism is a crucial method to alleviate the unseen-value problem. Furthermore, the copy mechanism can also guide the model to focus on dialogue context information. As shown in Table~\ref{tab:ablation}, we can see that the copy mechanism has a significant influence on the final performance.
\vspace{-2mm}
\paragraph{The effect of hierarchical encoder:}
To validate the effect of the hierarchical encoder, we concatenate all the dialogue utterances as a sequence and encode it using a bi-directional GRU. In the fine state decoder model, there is only a single decoder-encoder attention step. In this setting, we can see that there is the sharpest performance degradation. It indicates that the hierarchical encoding mechanism can provide more utterance information combined with the two-stage attention method.

\section{Conclusion}
In this work, we redefine the dialogue state into the structured state sequence. We leverage the coarse-to-fine generation to simplify the state generation process. With our well-designed model CREDIT, we get encouraging results, which is the best in BERT-free methods, in multi-domain DST tasks. Furthermore, we first try to use the reinforcement learning method to fine-tune the generative end-to-end DST model and get a performance improvement. In a word, we provide a tiny and efficient model for the DST community. For restricted-memory equipment, our CREDIT is a better choice with overall consideration.

\bibliographystyle{acl_natbib}
\bibliography{anthology,emnlp2020}

\appendix

\end{document}